\documentclass[conference]{IEEEtran}
\usepackage{blindtext, graphicx}
\usepackage[utf8]{inputenc}
\usepackage{amsmath}
\usepackage{color,soul}

\ifCLASSINFOpdf
\else
\fi
\begin{document}
\title{Authorship attribution via network motifs identification}

\author{\IEEEauthorblockN{Vanessa Queiroz Marinho$^{1,2}$, Graeme Hirst$^{1}$}
\IEEEauthorblockA{$^{1}$Department of Computer Science\\
University of Toronto\\
Toronto, ON, M5S 3G4, Canada\\
Email: vanessa9, gh@cs.toronto.edu}
\and
\IEEEauthorblockN{Diego Raphael Amancio$^{2}$}
\IEEEauthorblockA{$^{2}$Department of Computer Science\\
University of São Paulo\\
São Carlos, SP, 13566-590, Brazil\\
Email: diego@icmc.usp.br}
}


%Telephone: (800) 555--1212, Fax: (888) 555--1212}


\maketitle

\begin{abstract}
%\boldmath
Concepts and methods of complex networks can be used to analyse texts at their different complexity levels.
Examples of natural language processing (NLP) tasks studied via topological analysis of networks are keyword identification, automatic extractive summarization and authorship attribution. Even though a myriad of network measurements have been applied to study the authorship attribution problem, the use of motifs for text analysis has been restricted to a few works. The goal of this paper is to apply the concept of motifs, recurrent interconnection patterns, in the authorship attribution task. The absolute frequencies of all thirteen directed motifs with three nodes were extracted from the co-occurrence networks and used as classification features. The effectiveness of these features was verified with four machine learning methods. The results show that motifs are able to distinguish the writing style of different authors. In our best scenario, 57.5\% of the books were correctly classified. The chance baseline for this problem is 12.5\%. In addition, we have found that function words play an important role in these recurrent patterns. Taken together, our findings suggest that motifs should be further explored in other related linguistic tasks.
\end{abstract}


% \fi
\IEEEpeerreviewmaketitle

\section{Introduction}

The representation and characterization of real-world systems with complex networks has been useful to describe a large variety of systems found in the real world~\cite{Albert2002}. Some examples include the cell, which can be described as a network of substances linked by chemical reactions, and the Internet, a network of routers and computers connected by physical links~\cite{Mihalcea}. Traditionally, the study of networks was mostly limited to the analysis of random graphs. One of the pioneers in graph theory was the mathematician Leonard Euler who, in 1736, solved the famous
problem known as The K\"onigsberg Bridges~\cite{BarabasiLinked}. Since then, graph theory has benefited from major advances~\cite{boccaletti06}, mainly after the works of Watts and Strogatz~\cite{Watts-Colective-1998} and Barab\'asi and Albert~\cite{Barabasi99emergenceScaling} on the ubiquitous properties of real-world networks.

Mathematically, a network is an ordered pair $G = \{V, E\}$ formed by a set $V = \{v_1, v_2, ..., v_n\}$ of vertices (or nodes) and a set  $E = \{e_1, e_2, ..., e_m\}$ of edges. The network connectivity can be represented with an adjacency matrix $A$. In this matrix, the possible values for the element $A_{ij}$ are 0 or 1, where $A_{ij} = 1$ iff nodes $i$ and $j$ are connected. Networks can be directed or undirected, a property that depends on the reciprocity of the modelled system.  In addition, networks can be formed of weighted or unweighted edges~\cite{Costa2005}.  

In the last few years, the finding that many real systems could be characterized by networks with non-trivial patterns~\cite{Newman2010} allowed rapid development in the area. Some non-trivial patterns include the universal properties known as small-world~\cite{Watts-Colective-1998} and scale-free~\cite{Barabasi99emergenceScaling}. A rapid increase in data availability and computational capacity allowed the analysis and development of efficient algorithms in several applications, including text analysis via topological characterization of networks. There are several representations of texts as networks, where both nodes and edges may represent distinct textual aspects. In the most common models, nodes represent words and edges are established according to syntactic~\cite{Amancio2012b}, semantic~\cite{semantical} or empirical~\cite{empirica} relationships. Interestingly, it has been shown that the small-world and scale-free properties arise in such networks~\cite{Cancho2001}, in many cases as a consequence of Zipf's Law~\cite{zipf_1949}.

A special case of syntactic networks is the word co-occurrence network (or adjacency network)~\cite{Cancho2001}. In this type of network, links are established by connecting adjacent words, since most of the syntactical relations occur among neighbouring words~\cite{Cancho2004b}. The representation of texts as word co-occurrence networks has proven useful to tackle different tasks, for example, to create automatic extractive summarizers~\cite{Amancio2012b} and to identify the authorship of books~\cite{Amancio2011a,Mehri}.

Authorship attribution methods are relevant in practice because they can be applied to classify literary works and solve copyright disputes~\cite{Grant}. The first statistical authorship attribution techniques were devised by Mosteller and Wallace to probe the authorship of the so-called Federalist Papers~\cite{Mosteller}. Since then, many works tried to define characteristics to quantify the writing style of authors~\cite{Holmes}.%, known as stylometry.

Recent authorship attribution approaches based on networks~\cite{Amancio2011a,Mehri,Antiqueira2006a} quantify authors' writing style by using various measurements that characterize the topology of networks. In general, the topology of a complex network can also be characterized by the number of \emph{motifs} found on its structure. Motifs are small interconnection patterns (subgraphs) forming the topology of networks. Such structures have been able to characterize networks from different fields, as biochemistry, neurobiology, ecology, engineering and social networks~\cite{Milo,Kashtan,Milo2}. 
In most of these areas, specific network motifs have been found to play specific roles in the functional network activity. As a consequence, it has been shown that the prominence of particular subgraphs might be used to automatically identify the function of networks in many fields.
%
%From these findings, we believe that these patterns can be applied in different tasks. 
%
Although motifs have been used to detect universal patterns in different languages~\cite{Milo2} and to perform textual analysis~\cite{AmancioVoynich}, no comprehensive study has used these structures to characterize and distinguish different writing styles. 
In particular, we test the hypothesis that authors' writing preferences can be captured in an artlessly manner via identification of network motifs. As such, the main goal of this paper is to probe the relevance of network motifs as features for the authorship attribution problem.

%are interested in finding recurring patterns of a single author for the authorship attribution task. 
\section{Related Work}

In a typical authorship attribution problem, a text whose authorship is unknown is assigned to an author from a set of candidates. The first authorship attribution activities supported by statistical methods are from the XIX century. The goal of these methods is to maximize the probability for a text \textit{x} to belong to a possible author \textit{a}. Mosteller and Wallace~\cite{Mosteller} analysed the authorship of a collection of political essays, known as The Federalist Papers. With that study, they made a paramount contribution to the area showing that the frequency of common (function) words (as \emph{and} and \emph{to}) can distinguish different authors.

The research carried out by Mosteller and Wallace~\cite{Mosteller} was based mostly on simple textual statistics. After that seminal study, researchers have been proposing new attributes in order to characterize writing styles~\cite{Juola}. Some attributes traditionally used in the task include statistical properties of words (e.g., the average length, the frequency, burstiness and the vocabulary richness)~\cite{Stamatatos} and characters (frequencies and long-range correlations)~\cite{Grant}. In addition, syntactic (for example, frequency of specific chunks) and semantic attributes have been used as relevant attributes~\cite{Stamatatos,Hirst}. 

In the context of authorship attribution, Uzuner and Katz~\cite{Uzuner2005} used a corpus comprising $49$ books. They reported accuracies between 34\% and 87\%. Their best performance was achieved with the frequency of function words.  Grieve~\cite{Grieve} used 34 features and analysed their performance for different sets of authors. An accuracy of 80\% was reached with the frequency of words and punctuation marks using a set of 10 authors. Hirst and Feiguina~\cite{Hirst} combined the idea of bigram frequencies with syntactic analysis. Their syntactic label bigrams were found useful to distinguish the works of Anne Bront\"e and Charlotte Bront\"e, with an accuracy of 99.5\%. Jankowska et al.~\cite{Jankowska} observed that differences in the usage frequencies of the most common n-grams of characters and words could distinguish different authors. Sapkota et al.~\cite{Sapkota} demonstrated that character 3-grams that capture information about affixes and punctuation marks are important features for the authorship attribution task.

The dependencies between the topology of word co-occurrence networks and the writing style of authors were observed in some works~\cite{Amancio2011a,Mehri,Antiqueira2006a}. The measurements extracted from the networks by Antiqueira et al.~\cite{Antiqueira2006a} were able to distinguish different authors. After the combination of some measurements, it was also possible to cluster different authors, characterizing a common writing style. Amancio et al.~\cite{Amancio2011a} obtained an accuracy rate of 65\% when network measurements were combined with the intermittency of the distribution of words along the text. They found that most  topological measurements captures syntactic and stylistic characteristics of the language. Lastly Mehri, Darooneh and Shariati~\cite{Mehri} modelled 36 Persian books as co-occurrence networks to be used in the authorship attribution task. They did not employ any pre-processing step. The authorship was correctly assigned for 28 books, which represents an accuracy of 77.7\%.

In addition to some measurements in complex networks, motifs are structures that can be used to characterize the topology of co-occurrence networks. Motifs are recurrent  interconnection patterns found more frequently in real-world networks than in randomized ones~\cite{Milo}. These structures are represented by small subgraphs, generally involving three or four nodes. Milo et al.~\cite{Milo} used a variety of real-world networks (as such transcription networks, food webs and neuronal networks) and extracted the frequency of all possible subgraphs in those networks. The frequency of each subgraph was analysed and compared to its frequency in random graphs. As a consequence, the authors discovered the significant presence of motifs in all networks. As one of their conclusions, they pointed out a structural similarity between the neuronal connectivity and the transcription networks, because these networks presented similar frequencies of their motifs. Moreover, completely different frequencies of motifs were found in electronic circuits with distinct functionalities~\cite{Milo,Kashtan}.

Motifs were already used to analyse text~\cite{Milo2,AmancioVoynich}. Milo et al.~\cite{Milo2} analysed the frequency of motifs in word adjacency networks from texts in English, French, Spanish and Japanese. Despite the differences among these four languages, the authors discovered that they present very similar interconnection patterns. {One possible explanation for that is that languages possess an intrinsic structure, which divides words into categories}. Therefore, words from one category (e.g. prepositions) tend to be adjacent with others from different categories (e.g. nouns or articles)~\cite{Allen}. Amancio et al.~\cite{AmancioVoynich} realized that some motifs, as the ones labelled as 12 and 13 in Figure~\ref{motifs} below, rarely happen in real texts. 

\section{Methodology}\label{methodology}

\subsection{Dataset}
The dataset employed in this work comprises 40 novels, written by eight authors. The novels were published between 1835 and 1922. We obtained electronic versions of the
books from the Project Gutenberg repository\footnote{http://www.gutenberg.org}. The full list with the $40$ books is summarized in Appendix~\ref{books}. To avoid the influence of different text lengths in the frequency of motifs and network measurements, each novel was truncated to the length of the shortest book in each scenario used in this paper.

\subsection{Pre-processing of the text}\label{preprocessing}
The task of modelling text as a complex network can be divided into two steps, the pre-processing of the text
and the connection of words. One of the first pre-processing activities is to remove punctuation marks.
In this paper, contractions (as \emph{he's} and \emph{isn't}) were kept. We believe that the choice of writing \emph{isn't} instead of its equivalent form \emph{is not} is related to the writing style and, therefore, this should be mapped differently in the network. 

One of the pre-processing steps usually applied in text analysis via networked models is the removal of function words or stopwords. These words are mainly prepositions, articles and pronouns that convey little semantic content. The set of stopwords removed in some of our scenarios is presented in Appendix~\ref{stopwords}. However, these words will be included in the creation of the network in some of our experiments. This is based on the fact that these words are very useful in traditional techniques of authorship attribution~\cite{Mosteller}.

The lemmatization of the text is another pre-processing step that can be applied. It is performed with a part-of-speech tagger. In this step, plural words are changed to their singular version, verbs to their infinitive form and names to their masculine form. After this step, words with the same lemma are mapped into the same node. The part-of-speech tagger used to perform the lemmatization process is described in~\cite{Bird}.

To demonstrate the two different pre-processing steps that will be used in this paper, Table~\ref{preprocessingTable} illustrates their applications in an extract of the book \emph{The Adventures of Sherlock Holmes}, by Arthur Conan Doyle. Here, these two steps will be combined and evaluated in order to find the best scenario for the authorship attribution task.

\begin{table*}[!t]
\caption{Example of the two pre-processing steps applied to sentences from the book \emph{The Adventures of Sherlock Holmes}.}
\label{preprocessingTable}
\centering
\begin{tabular}{l*{4}{l}r}
\hline
\textbf{Original extract}   & \textbf{Without stopwords + lemmatization}  \\
\hline
{``There are three men waiting for him at}   	& {three men wait} \\
{the door", said Holmes. ``Oh, indeed!}   		& {door say holmes oh indeed} \\
{You seem to have done the thing very} 	& {seem do thing} \\
{completely. I must compliment you."}  	& {completely must complement} \\
{``And I you", Holmes answered.}  	& {holmes answer} \\
\hline
\end{tabular}
\end{table*}

\subsection{Co-occurrence networks}

\begin{figure}[!t]
\centering
\includegraphics[width=1.9in]{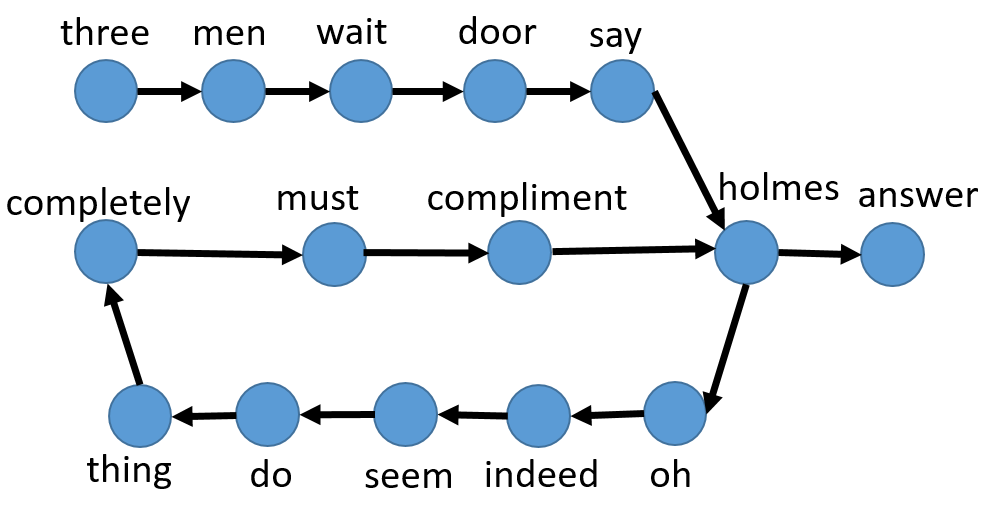}
\caption{The graph represents the directed co-occurrence network for the pre-processed sentence ``three men wait door say holmes oh indeed seem do thing completely must compliment holmes answer". This network was created connecting each word to its adjacent word and the direction of the edges is defined from the word in the left to the one in the right.}
\label{network}
\end{figure}

After the pre-processing steps, we need to connect each word to create the network. The written language is formed by linear chains of words; therefore, the easiest way to represent it 
is to connect adjacent words. This type of network, known as a co-occurrence or adjacency network, is 
widely used in the literature~\cite{Cancho2001,Amancio2011a,Roxas}. In a word co-occurrence network,
words are mapped into nodes, and a link between two nodes is established if the corresponding words appear adjacent at least once in the (pre-processed) text. The directed network that represents the pre-processed sentences of Table~\ref{preprocessingTable} is illustrated in Figure~\ref{network}.

%\I{-1cm}

\subsection{Measurements}\label{measurements}
\begin{figure}[!t]
\centering
\includegraphics[width=2.8in]{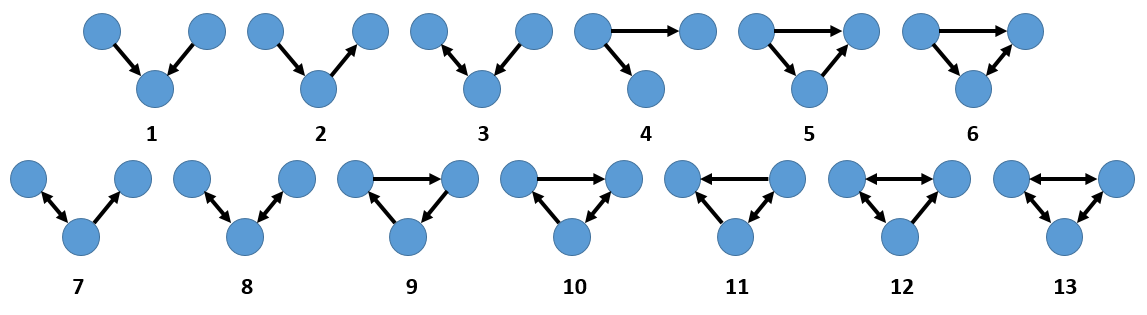}
\caption{All thirteen directed motifs involving three nodes.}
\label{motifs}
\end{figure}

The topological structure of a complex network can be characterized by several metrics. Firstly, the directed networks derived from our dataset will be characterized by the absolute frequency of all directed motifs involving three nodes. We did not employ motifs involving more nodes because we aimed to use
a small set of attributes.
The set of directed motifs with three nodes is presented in Figure~\ref{motifs}. The frequencies were extracted with the software \emph{mfinder}~\cite{mfinder}.

In order to compare the classification results obtained with the frequency of motifs and the ones when other networked measurements are used, we extracted five measurements from complex networks. We modified the directed network to obtain its undirected version. This was easily achieved by transforming the adjacency matrix $A$ into its symmetric form. Each one of the measurements are described below:
\begin{itemize}

\item Average Degree of Neighbors ($ADN_i$): The degree indicates the amount of distinct neighbors. The $ADN_i$ quantifies the average degree of all neighbors of $i$~\cite{pastor2001dynamical}.

\item Average Shortest Path Length ($L_i$): This measurement represents the average distance between node $i$ and all other nodes. In textual networks, $L_i$ quantifies the relevance of each word $i$ according to its distance to the most frequent words~\cite{Amancio2011a}.

\item Betweenness Centrality ($B_i$): The betweenness centrality  measures the relevance of a node $i$ according to the number of shortest paths that include this node~\cite{boccaletti06}. In textual networks, frequent words tend to have a high value for this measurement. In addition, some words may act as \emph{bridges} (or articulation points) connecting concepts from different communities and thus they also may take high values of betweenness, regardless of their frequency. Therefore, this measurement may quantify the diversity of contexts in which a word can be used~\cite{Amancio2011a}.

\item Clustering Coefficient ($CC_i$): The clustering coefficient of a node $i$ indicates the probability of two of its neighbors being connected. As it happens in many small-world networks, Cancho and Solé~\cite{Cancho2001} found that the clustering coefficient in textual networks is much higher than the expected value in equivalent random networks.

\item Assortativity ($r$): In some networks, nodes of certain degree tend to connect with similar nodes~\cite{Newman2002}. This measurement quantifies the degree correlation and it can be calculated with the Pearson correlation coefficient, $r$~\cite{Costa2007}. In most of the cases, co-occurrence networks are disassortative, $r < 0$.

\end{itemize}

\subsection{Extracting properties of books from properties of words}

Apart from assortativity and motifs, all measurements are locally defined, i.e. a value is assigned for each word. The goal is to obtain values that can be used as global measurements of each book, not only of each single word. Therefore, we adopted a summarization procedure. For each local measurement $X$, the most natural choice is to calculate the average $\left \langle X \right \rangle$, i.e. the average of $X$ over all the $M$ unique words. We also computed the deviation and skewness of the distribution, denoted as $\sigma(X)$ and $\gamma(X)$, respectively.  Note that this type of summarization has been performed in related studies~\cite{Amancio15b}.
In summary, the three features we use for each measurement $X = \{ADN, L, B, CC\}$ are:

\begin{equation}
\text{Average:} \hspace{1cm}  \left \langle X \right \rangle = \frac{1}{M} \sum_{i=1}^{M} X_i, \nonumber
\end{equation}
\begin{equation}
\text{Deviation:} \hspace{1cm}  \sigma(X) = \Bigg{[} \frac{1}{M-1} { \sum_{i=1}^{M}(X_i - \left \langle X \right \rangle)^{2} } \Bigg{]}^{1/2}, \nonumber
\end{equation}
\begin{equation}
\text{Skewness:} \hspace{1cm} \gamma(X) =
\left \langle \Bigg( \frac{X - \left \langle X \right \rangle}{\sigma(X)} \Bigg)^{3} \right \rangle. \nonumber
\end{equation}

\subsection{Machine Learning Methods}\label{patternMethods}

In order to quantify the ability of the motifs to distinguish among authors, we employed four machine learning algorithms to induce classifiers from a training set. The techniques are \emph{Support vector machines}, kNN, \emph{Naive Bayes} and C4.5~\cite{Duda}. We used the default configuration of these methods available in Weka~\cite{weka}. It has been shown that the default configuration of parameters in Weka provides near optimal performance for most of the cases~\cite{comparingPatterns}. These methods were applied to a training set independent of the test set using the cross validation technique with 10 folds~\cite{comparingPatterns}. In this technique, at each cycle one tenth of the books are used as test while the other nine tenths are used for training the algorithm.

\section{Results}\label{results}
As we described in Section~\ref{preprocessing}, different pre-processing steps may be applied before creating the network of each text. For each one of our experiments, we considered four different input scenarios: (i) original text, (ii) without stopwords, (iii) after the lemmatization process and (iv) after the removal of stopwords and the lemmatization process. For each input scenario, we derived one network for each book. In the first experiment, the absolute frequency of all thirteen motifs was extracted. Finding the set of motifs of size 3 is computationally feasible. The average time to extract all motifs in scenarios (ii) and (iv) was around 30 seconds per book. The derived networks in scenarios (i) and (iii) have more nodes and edges and, therefore, the time to extract the motifs increased to an average of 4 minutes per book. We applied the machine learning methods described in Section~\ref{patternMethods}. The results are presented in Table~\ref{motifResultsDirected}. In these results, the function words play an important role during the extraction of the motifs because the overall performance drops to very low values when such very frequent words are disregarded.

\begin{table*}[!t]
\caption{Percentage of books correctly classified when the absolute frequency of directed motifs was used as the only classification feature}\label{motifResultsDirected}
\centering
\begin{tabular}{l|l|l|l|l}
 \hline
 & \textbf{C4.5} & \textbf{kNN} & \textbf{SVM} & \textbf{Bayes}\\
 \hline
{(i) Original text}	 &	40.0\% & 55.0\% & 45.0\% &	45.0\%\\ \hline
{(ii) Without stopwords}	 & 	 27.5\%	&32.5\%	&0.0\%	&30.0\%\\ \hline
{(iii) Lemmatization process}	 &	57.5\%	& 45.0\% &	45.0\%	& 52.5\%\\ \hline
{(iv) Without stopwords + lemmatization}	 & 	 22.5\% &	27.5\% &	2.5\% &	30.0\%\\ \hline
\end{tabular}
\end{table*}

%\{-0.75cm}

As a way to evaluate whether the motifs are actually extracting a real pattern, a random verification step was performed. Only the input scenario (iii) was selected, since it led to the best results. In this process, each instance keeps the same frequencies as the original one; however, the class $y$ is randomly selected from the set of all possible authors. Note that the correct author is also included in the set. This process was performed 10 times and the averages of books correctly classified are 12\% (C4.5), 9\% (kNN), 13\% (SVM) and 10.8\% (Naive Bayes). These results are similar to the chance baseline for this problem, 12.5\%, since each one of the 8 authors has the same probability of being randomly selected. Therefore, these results and the ones presented in Table~\ref{motifResultsDirected} confirm that the frequency of motifs is able to significantly extract patterns related to the authorship of each book. 

\begin{figure*}[!t]
\centering
\includegraphics[width=5.1in]{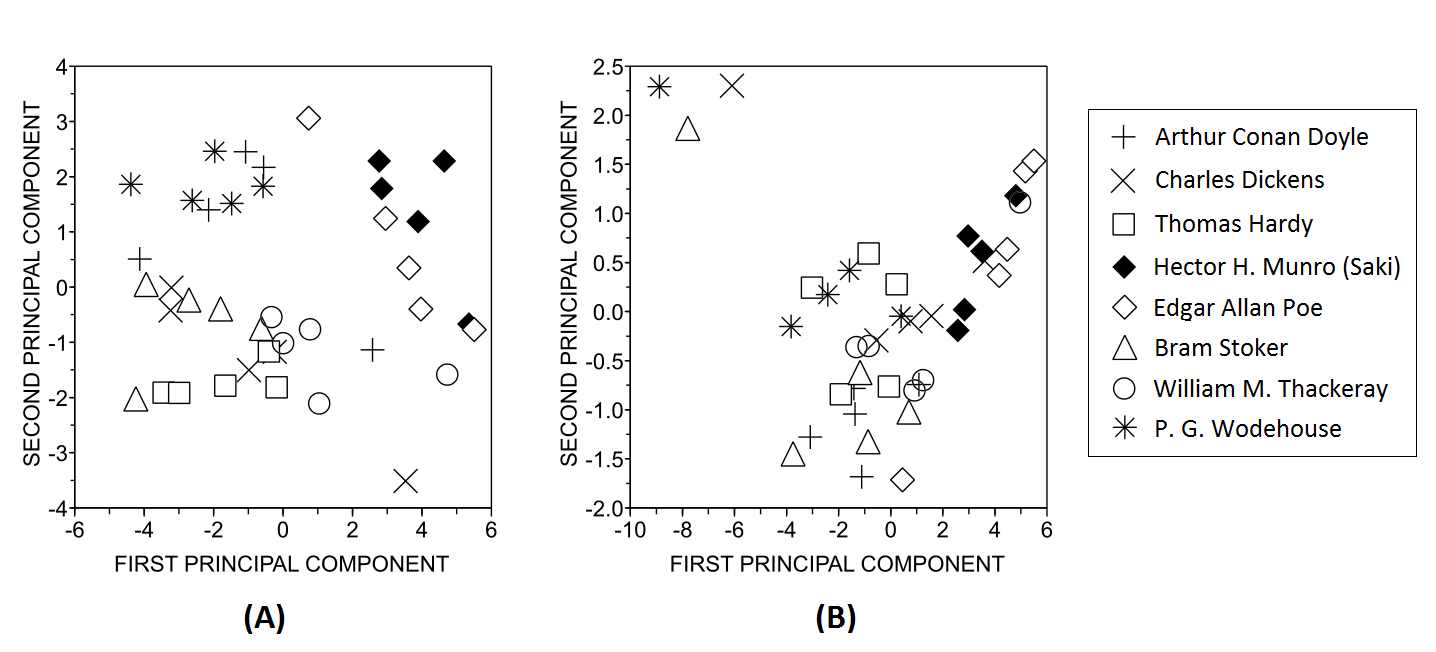}
\caption{Principal component analysis performed using network motifs as attributes. The input scenarios were: (A) Scenario (i) - original text; and (B) Scenario (ii) - text without stopwords. Note that the discriminability increases when the stopwords are considered.}
\label{pca}
\end{figure*}

A visualization of the classification for two different input scenarios is provided in Figure~\ref{pca}. Each one of the 8 authors is represented by a different symbol and each element in the plot is a distinct book. The highest accuracy rate found for scenario (i) was 55\%. The ability to distinguish among different authors is more evident in Figure~\ref{pca} (A), which represents scenario (i). The lowest accuracy rates were found for scenario (ii), which is presented in Figure~\ref{pca} (B). This poor discriminability of authors can be attributed to the fact that stopwords were disregarded in that scenario.

In our last experiment, several features were extracted from the books. Firstly, the network measurements explained in Section~\ref{measurements} were extracted from the undirected networks of each book. The results for the four scenarios are listed in Table~\ref{networkResults}. Besides that, the relative frequency of the 20 most frequent words in all 40 books were extracted from the input scenario (i) and used as a classification feature. This result is presented in the last row of Table~\ref{networkResults}. Comparing the results of Tables~\ref{motifResultsDirected} and~\ref{networkResults}, the motif extraction for the scenarios (i) and (iii) performed almost as well as the results retrieved with traditional network measurements. However, some complex network measurements are correlated with the frequency of words, e.g. betweenness centrality, and this may improve the performance of such measurements. Moreover, the network measurements employed in this paper are not so sensitive to the removal of the function words. The results obtained with the frequency of common words were relatively high. The 20 most frequent words were mainly function words (e.g. \emph{and}, \emph{he}, \emph{his}). These results were already expected, since function words can be used as a strong tool to detect writing styles. However, such features are prone to manipulation, a disadvantage not present in networked approaches. 
\begin{table*}[!t]
\caption{Percentage of books correctly classified when several features were extracted from the books}\label{networkResults}
\centering
\begin{tabular}{l|l|l|l|l}
 \hline
 & \textbf{C4.5} & \textbf{kNN} & \textbf{SVM} & \textbf{Bayes}\\
 \hline
{(i) Original text}	 &	  50.0\% & 42.5\% & 42.5\% & 55.0\%\\ \hline
{(ii) Without stopwords}	 & 	 37.5\%	&45.0\%	&27.5\%	&37.5\%\\ \hline
{(iii) Lemmatization process}	 &	  47.5\% & 50.0\%	& 37.5\% &	45.0\%\\ \hline
{(iv) Without stopwords + lemmatization}	 & 32.5\% & 37.5\%	& 32.5\% & 40.0\%\\ \hline
\hline
{Frequency of 20 most frequent words} & 55.0\% & 67.5\% &  72.5\% & 55.0\%\\
\end{tabular}
\end{table*}

\section{Conclusion}\label{conclusion}
The authorship attribution task has been studied with some success through the representation of word co-occurrence networks. In this paper, we investigated
whether network motifs found on these networks are able to discriminate different authors.

The accuracies obtained in this paper are relevant, which confirms that motifs are able to capture aspects of the writing style of different authors. Besides that, the results are considerably higher than the chance baseline expected for this problem, 12.5\%. Therefore, we can conclude that there is a dependency between the frequency of the motifs and the writing style of different authors. Another interesting result is the fact that function words played a crucial role to the success of our approach. When the motifs extraction was performed in networks without those words, the accuracies decreased and even reached a level that one of the classifiers could not find any pattern to perform the classification.

The focus of this paper was to evaluate the accuracies when motifs were used as the only classification attributes. To our knowledge, this is the first time that network motifs are used to identify authorship in written texts. Our best result, 57.5\% is lower than some traditional approaches found in the literature, that are around 90\%~\cite{Hirst,Uzuner2005,Grieve}. Comparing our results with others that use co-occurrence networks, Amancio et al.~\cite{Amancio2011a} obtained an accuracy rate of 65\%. From the dataset used by Mehri, Darooneh and Shariati~\cite{Mehri}, 28 books (77.7\%) were correctly classified. The results retrieved in this paper may be seen as complementary measurements and can be combined with traditional techniques usually employed in authorship attribution. Moreover, the significative results obtained here suggest that the characterization based on motifs should be further explored in other related linguistic tasks, such as language identification or complexity in authors' writing style.

\appendices
\section{List of books used in our experiments}\label{books}
The list of 40 books is presented in Table~\ref{tableBooks}.
\begin{table}[!t]
\renewcommand{\arraystretch}{1.3}
\caption{List of books employed in the authorship attribution task.}
\label{tableBooks}
\centering
\begin{tabular}{l|p{5cm}}
\hline
 \textbf{Author} & \textbf{Books} \\
\hline
Arthur Conan Doyle	 & The Adventures of Sherlock Holmes (1892), The Tragedy of the Korosko (1897), The Valley of Fear (1914), Through the Magic Door (1907), Uncle Bernac - A Memory of the Empire (1896).\\
\hline
Bram Stoker	& Dracula’s Guest (1914), Lair of the White Worm (1911), The Jewel Of Seven Stars (1903), The Man (1905), The Mystery of the sea (1902).\\  
\hline
Charles Dickens 	& A Tale of Two Cities (1859), American Notes (1842), Barnaby Rudge: A Tale of the Riots of Eighty (1841), Great Expectations (1861), Hard Times (1854).\\
\hline
Edgar Allan Poe	 & The Works of Edgar Allan Poe, Volume 1 - 5, (1835). \\ 
\hline
Hector H. Munro (Saki)	 & Beasts and Super Beasts (1914), The Chronicles of Clovis (1912), The Toys of Peace (1919), When William Came (1913), The Unbearable Bassington (1912).\\
\hline
P. G. Wodehouse
& Girl on the Boat (1920), My Man Jeeves (1919), Something New (1915), The Adventures of Sally (1922), The Clicking of Cuthbert (1922).\\
\hline
Thomas Hardy 	& A Pair of Blue Eyes (1873), Far from the Madding Crowd (1874), Jude the Obscure (1895), Mayor Casterbridge (1886), The Hand of Ethelberta (1875). \\
\hline 
William M. Thackeray  & Barry Lyndon (1844), The Book of Snobs (1848), The History of Pendennis (1848), The Virginians (1859), Vanity Fair (1848).\\  
\hline
\end{tabular}
\end{table}

\section{List of stopwords}\label{stopwords}
\emph{i, me, my, myself, we, our, ours, ourselves, you, your, yours,
yourself, yourselves, he, him, his, himself, she, her, hers,
herself, it, its, itself, they, them, their, theirs, themselves,
what, which, who, whom, this, that, these, those, am, is, are,
was, were, be, been, being, have, has, had, having, do, does,
did, doing, a, an, the, and, but, if, or, because, as, until,
while, of, at, by, for, with, about, against, between, into,
through, during, before, after, above, below, to, from, up, down,
in, out, on, off, over, under, again, further, then, once, here,
there, when, where, why, how, all, any, both, each, few, more,
most, other, some, such, no, nor, not, only, own, same, so,
than, too, very, s, t, can, will, just, don, should, now}

\section*{Acknowledgment}
V.Q.M. and D.R.A. acknowledge financial support from FAPESP (grant nos. 14/20830-0, 15/05676-8 and 15/23803-7). G.H. acknowledges The Natural Sciences and Engineering Research Council of Canada.
 
%This work was financially supported by grants , São Paulo Research Foundation (FAPESP).

\bibliographystyle{IEEEtran}


%\bibliography{IEEEabrv,references} 

\end{document}